\documentclass[10pt,twocolumn,letterpaper]{article}

\usepackage{cvpr}
\usepackage{times}
\usepackage{epsfig}
\usepackage{graphicx}
\usepackage{amsmath}
\usepackage{amssymb}

\usepackage[breaklinks=true,bookmarks=false]{hyperref}

\cvprfinalcopy 

\def\cvprPaperID{****} 

\begin{document}

\title{Reconstructing Network Inputs with Additive Perturbation Signatures}

\author{Nick Moran, Chiraag Juvekar\\
Algorithmic Systems Group, Analog Devices\\
125 Summer Street, Boston MA\\
{\tt\small nicholas.moran@analog.com, chiraag.juvekar@analog.com}
}

\maketitle
\section{Introduction}

In recent years, deep neural networks have delivered state-of-the-art results on a wide variety of tasks in computer vision, natural language processing and many other subfields.  At the same time, the models driving this success have become larger and more complex, requiring ever-growing computing power to train and run.  As a result, we have seen the rise of Machine Learning as a Service (MLaaS)\cite{chappell2015introducing}, in which models are trained and evaluated by third parties rather than the end user.

This trend presents significant privacy concerns with regards to both the training data used to build models, as well as the input data at inference time.  Given various levels of access to a trained model, its parameters, and/or its outputs, how much information can an adversary recover about its training data or its inputs?

In this work, we present preliminary results demonstrating the ability to recover a significant amount of information about secret model inputs given only very limited access to model outputs and the ability evaluate the model on additive perturbations to the input.

\section{Background}

Significant work has been done on understanding and mitigating the ability of an adversary to recover training data given acess to the trained model \cite{papernot2016semi}\cite{shokri2017membership}.  Less work has been done on the related problem of recovering inference inputs given access to some subset of model outputs.  Most similar to our setting is the method presented in \cite{yang2019adversarial}, in which model inputs are recovered given access to a truncated set of class logits output by the model.

Homomorphic encryption has been proposed as a way to allow an untrusted third party to perform inference on sensitive data in a way that limits disclosure of inputs \cite{gilad2016cryptonets}.  In this setting, the model owner receives only the encrypted inputs, and may also receive only a limited amount of information about the network outputs.  In this work, we investigate whether even this limited information is sufficient to recover sensitive inputs.

\begin{figure}
    \centering
    \includegraphics[width=\columnwidth]{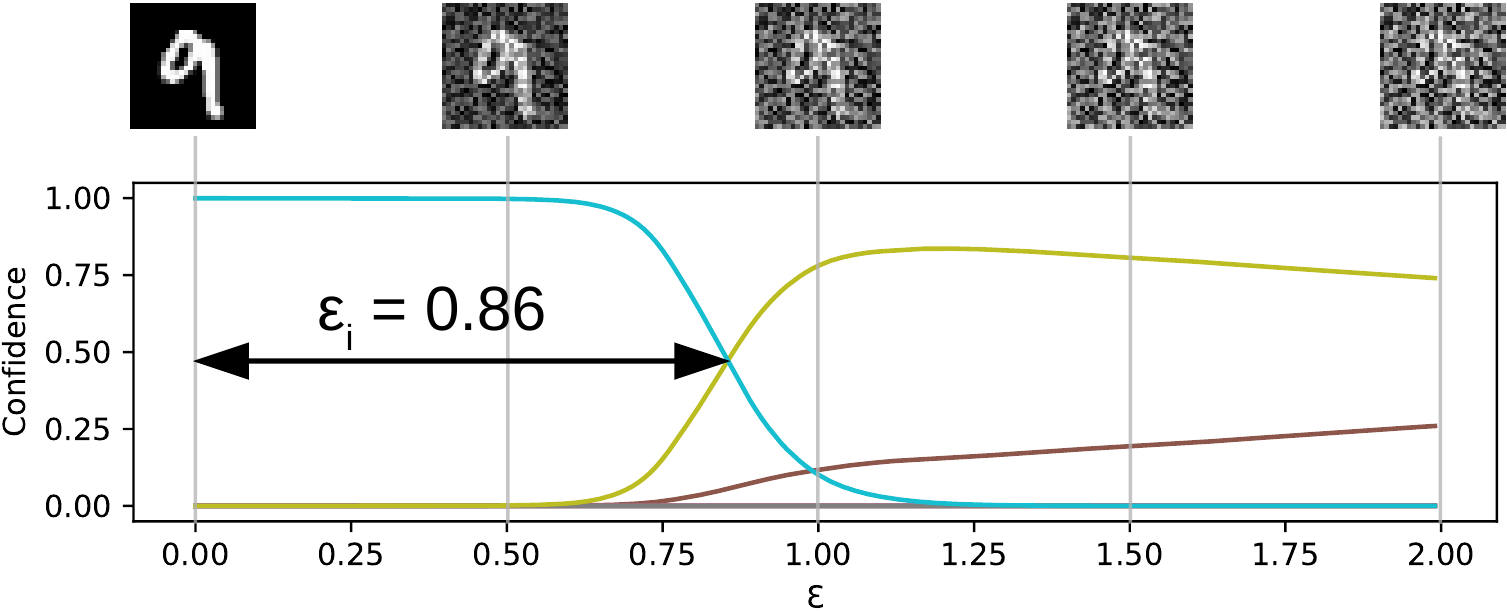}
    \caption{The method for calculating $\epsilon_i$ for a particular noise direction.  The chart shows how class confidences change as $\epsilon$ increases.  The point at which the original class is no longer the most probable defines the value of $\epsilon_i$.}
    \label{fig:epsilon}
\end{figure}

\section{Attack Setting \& Notation}

We consider the following scenario.  A model $M$ is trained to perform a classification task and made available to users.  In this work we will specifically consider a $M$ to be a convolutional neural network (CNN), and will present results from the MNIST\cite{lecun1998mnist} classification task as a simple baseline.  A user has a sensitive input, $X$ (we will use $x$ to represent other non-sensitive inputs), on which they would like to evaluate $M$.  We will denote by $M(X)$ the identity of the model's top-1 prediction for $X$.  Inference is performed under encryption such that $X$ is kept secret even from the owner of $M$.  An adversary attempts to recover $X$ given access to the following capabilities.  The adversary (who may be the same as the model owner) has black-box access to evaluate $M$ on their own inputs, and is further capable of inducing the evaluation of $M$ on additive perturbations of $X$, $\hat{X} = X + N$, where $N$ is chosen by the adversary, and can detect \textit{only} whether $M(X) \stackrel{?}{=} M(\hat{X})$.  In other words, the adversary does not know the actual value of the most-probable class, neither for $X$ nor $\hat{X}$.  The goal of the adversary is to recover as much information as possible about $X$.

This scenario is inspired by the setting of inference under homomorphic encryption, in which the additive homomorphic property allows the injection of additive perturbations to the input without exposing the initial value of the input.  In such a setting, the model's final output may similarly be encrypted, but changes in the top class may be exposed by changes in some down-stream behavior.  We note also that in many settings, an adversary may well have access to the final output, and we should expect that access to such information will further enhance the adversary's capabilities.

\section{Method}

Our method works by measuring the sensitivity of the model to additive perturbations to $X$ in a set of fixed directions, and training a decoder model to predict the pixel-values of $X$ given these sensitivities.  We begin by selecting a set (in this work we use a set of size 100) of random perturbation directions, which we will term $N = \{N_i\}$.  Each perturbation direction is a real-valued tensor of the same shape as $X$; in the case of MNIST they are of shape $28\times28$.

For a given input image, $x$, we compute a set of minimal $\epsilon_i$ such that $M(x + \epsilon_i \cdot N_i) \neq M(x)$ and $\epsilon_i > 0$.  This set forms a fingerprint vector $E_x$ = [$\epsilon_0$, $\epsilon_1$,...] which is the input to our decoder.   This process is shown in Figure~\ref{fig:epsilon}.

To train our decoder, we build a training set of ($E_x$,$x$) pairs.  In this work, we will assume for simplicity that the adversary has access to the same training set as was used to train $M$, although this need not be the case.  We will use a variant of the generator architecture from \cite{radford2015unsupervised}, in which the input vector is projected into a tensor with small spatial extent, which is then upsampled and refined into a full-resolution image using convolution transpose layers.  The decoder is trained to minimize the $L_2$ error between the ground truth $x$ and its reconstructions from $E_x$.

\section{Results}

\begin{figure}
    \centering
    \includegraphics[width=\columnwidth]{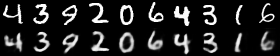}
    \caption{Reconstructions (bottom row) of held-out test data (top row) from 100-element signatures.}
    \label{fig:reconstructions}
\end{figure}

Figure~\ref{fig:reconstructions} shows the results of our method on images from the MNIST test set.  We observe first that the decoder is clearly able to recover the correct digit class of each image, despite receiving only $E_x$ as input with no information about class identities.  Further, the decoder is able to recover a significant amount of stylistic information such as digit angle.

\begin{figure}
    \centering
    \includegraphics[width=\columnwidth]{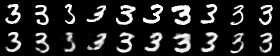}
    \caption{Reconstructions (bottom row) of a single digit type and corresponding original digits (top row).}
    \label{fig:threes}
\end{figure}

Figure~\ref{fig:threes} shows the results of our method applied to a variety of samples of a single digit type.  These results more clearly demonstrate the ability of our method to recover additional information beyond the image class alone.  Differences in orientation, stroke width, and curvature are all apparent in the recovered images.

\section{Future Work}

Although the MNIST dataset is convenient for rapid prototyping, it also may present a significantly easier task than other, more realistic datasets.  On larger datasets, it may not be possible to perform pixel-level recovery of inputs.  Instead, we hope to recover other, lower-dimensional attributes of the data. 

We currently select the perturbation directions uniformly at random.  However, we hypothesize that more cleverly-chosen perturbations may produce more informative signals for the decoder.  In particular, we intend to explore directions derived from adversarial attacks\cite{szegedy2013intriguing}, as well as directions obtained by subtracting the centroids of training classes.

Finally, we are exploring possible defenses to this attack, either by training models which are robust to this sort of inversion, or by altering inputs to fool the decoder.

{\small
\bibliographystyle{ieee_fullname}
\bibliography{reconstructing}
}

\end{document}